\documentclass[10pt,twocolumn,a4paper]{esaAI}

\title{Operational range bounding of spectroscopy models\\with anomaly detection}

\def\authorEmail{luis.simoes@mlanalytics.pt}

\author[1]{Lu{\'i}s F. Sim{\~o}es\thanks{Corresponding author. E-Mail: \authorEmail}}
\author[2]{Pierluigi Casale}
\author[1]{Mar{\'i}lia Felismino}
\author[3]{\\Kai Hou Yip}
\author[3]{Ingo P. Waldmann}
\author[3]{Giovanna Tinetti}
\author[4]{Theresa Lueftinger}
\affil[1]{ML Analytics, Lisbon, Portugal}
\affil[2]{OPIT - Open Institute of Technology, St. Julian's, Malta}
\affil[3]{Department of Physics and Astronomy, University College London, London, UK} % Gower Street, London, WC1E 6BT [removing full address for consistency]
\affil[4]{European Space Agency, ESA-ESTEC, Noordwijk, The Netherlands}
\begin{document}

% Creates the title and author list
\makeCustomtitle

\begin{abstract}
Safe operation of machine learning models requires architectures that explicitly delimit their operational ranges.
We evaluate the ability of anomaly detection algorithms to provide indicators correlated with degraded model performance.
By placing acceptance thresholds over such indicators, hard boundaries are formed that define the model's coverage.
As a use case, we consider the extraction of exoplanetary spectra from transit light curves, specifically within the context of ESA's upcoming Ariel mission.
Isolation Forests are shown to effectively identify contexts where prediction models are likely to fail.
Coverage/error trade-offs are evaluated under conditions of data and concept drift.
The best performance is seen when Isolation Forests model projections of the prediction model's explainability SHAP values.
\end{abstract}

\section{Introduction}

As Machine Learning adoption grows across risk-averse industries such as aerospace, various efforts are underway to design the validation processes that will ensure safety in complex operational settings \cite{tambon2022certify, mamalet:hal-03176080, ECSS:MLQHandbook, MLEAP_interim_technical_report, Borg2019}.

\vspace{5pt}
\noindent
\textbf{The safety cage architecture} being advocated by such efforts consists of a safety mechanism that continuously monitors system observables, and intervenes if it judges the model will exceed the bounds of a safe domain for a particular input.
Traditional monitoring systems run asynchronously from the prediction model and perform evaluations over assembled production data \cite{treveil2020introducing}. The safety cage architecture, instead, runs a monitoring system in parallel with the prediction model, and operates at an individual sample level.
The acceptance or rejection of individual samples effectively places boundaries across the model's input space, thus explicitly defining its operational range.
Safety cages must have ways to detect and address different kinds of failure modes. The current work focuses on issues of performance degradation due to differences between training and production data.

% - The approach followed here addresses the identification of epistemic uncertainty (lack of training data such as that being encountered post-training). However, it should be noted that a model's performance in a region beyond the training set is not guaranteed to be incorrect (the model may properly generalize there), and also, there will be regions within the training data where the model obtains poor performance.

% - hazards of addressing a model's error by building another one, over the same dataset, that makes its own, distinct errors. Recommendation in \cite{ECSS:MLQHandbook}: "Do not mitigate ML with more ML".

% "As a best practice for the ML training, as soon as the rules driving the safety cage are defined, they should direct the learning of the ML algorithm" \cite{ECSS:MLQHandbook}

\vspace{5pt}
\noindent
\textbf{Exoplanet transit spectroscopy.}
Atmospheres of exoplanets are studied using transit spectroscopy. As a planet passes through the projected surface of the host star, the observed light curve will display a temporary reduction (a dip) in the overall flux from the planetary system.
% (\cref{fig:light_curves})
The level of reduction varies in different wavelength channels, which is key to revealing the chemical and dynamical properties of the planet's atmosphere. These changes are typically 100 ppm or less in magnitude, making them easily buried under different instrumental and astrophysical noises orders of magnitude higher than the signal.

The Ariel Mission is a European Space Agency (ESA) M4 mission that will conduct the first comprehensive study of 1000 exoplanets in our galactic neighbourhood \cite{tinetti2018chemical, tinetti2021ariel}.
The Ariel Data Challenge \cite{ADC2019paper, ADC22:ABC, ADC22:lessons, ADC24:Kaggle} is a series of data competitions organised by mission scientists and engineers within the Ariel Consortium to invite innovative solutions to challenging problems faced by the ESA Ariel mission, and by extension, the Planetary Science community.
The 2019 and 2021 editions of the challenge \cite{ADC2019paper} looked for innovative detrending solutions to extract planetary spectra from observed spectroscopic light curves that are contaminated by the instrument's systematic noise effects,
% (such as ramps),
as well as stellar activities from the host stars. The dataset simulations represent Ariel observations with mission-dedicated radiometric models to ensure good representation of the instrument. The simulations can however never represent the actual data, which will suffer from unknown unknowns.
Ensuring the reliable operation of Machine Learning models, as they move from simulated to real data, with unknowable ground truths, is of paramount importance.
% This motivates the research presented here, into the effectiveness of safety cage architectures.

\begin{figure*}[t]
    \centering
    \includegraphics[width=\textwidth]{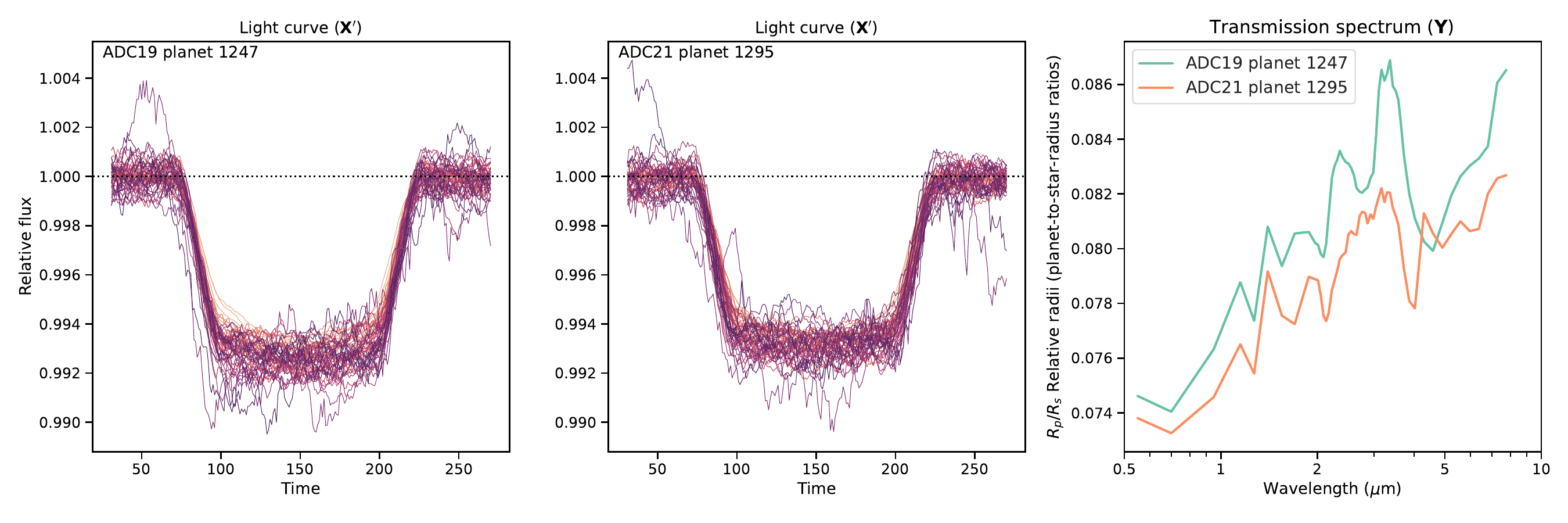}
	\caption{Transit light curves for the same star/planet system, produced by the simulation pipelines
    %\cite{morello2020exotethys, waldmann2015tau, mugnai2020arielrad, sarkar2021exosim}
    of the ADC19 \cite{morello2020exotethys, waldmann2015tau} (left) and ADC21 \cite{mugnai2020arielrad, sarkar2021exosim} (middle) challenges (seen: 55 light curves, one per wavelength channel, transformed through a sliding-window aggregation over time and across the 10 photon noise instances of the star's first stellar spot configuration). Right plot: the planet's transmission spectrum (55 channels) the trained model needs to predict from the 55 light curves.
    Simulations for exoplanet Qatar-4b, using star/planet parameters from the Ariel Mission Candidate Sample \cite{MCS:edwards2019updated,MCS:edwards2022ariel}.}
    % MCS match found against https://github.com/arielmission-space/Mission_Candidate_Sample/blob/main/target_lists/Ariel_MCS_Known_2024-03-27.csv
    % Qatar-4 b -- https://exodb.space/exoplanet/4016 -- https://en.wikipedia.org/wiki/Qatar-4b
    % ^ The Qatar-4 match is the closest one overall we have between the ADC19/21 datasets and the MCS csv.
    % ^ The ADC19 star/planet parameters come from https://arxiv.org/abs/1905.04959
    %   Table 6 has an exact match for Qatar-4b [same star_temp & star_k_mag]
	\label{fig:light_curves}
\end{figure*}

\section{Results}

\begin{figure*}[t]
    \centering
    \includegraphics[width=\textwidth]{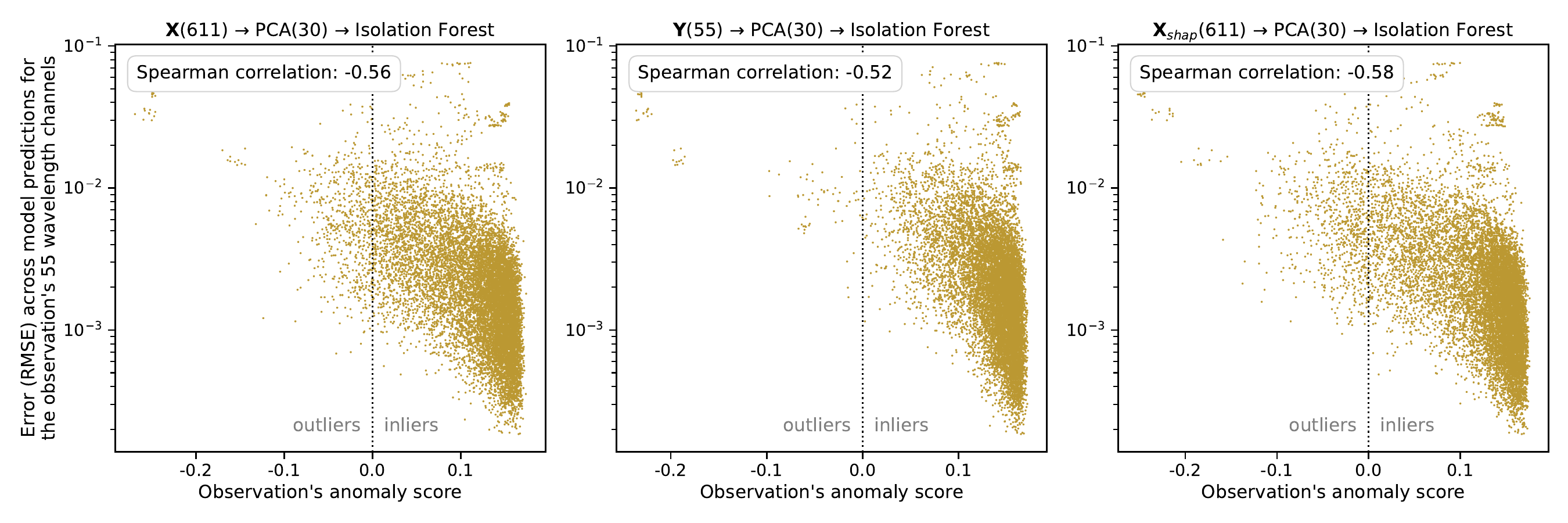}
	\caption{
    Relation between anomaly scores produced by Isolation Forests and predictive performance of another model trained over the same data.
    Results obtained through 10-fold cross-validation over the ADC21 dataset.
    Plots show values measured across the validation-folds for the whole dataset, from Isolation Forest, PCA and Ridge models trained over the respective train-folds.
    In the left plot, the Isolation Forest models the distribution of $\mathbf{X}$, the 611-dimensional representations of light curves that the Ridge model takes as inputs. In the central plot the modelled distribution is that of $\mathbf{Y}_{true}$, the spectra that Ridge models are trained to predict. In the right plot the Isolation Forest models the $\mathbf{X}_{shap}$ values produced by the SHAP Explainable AI library \cite{SHAP} that represent how the Ridge model transforms $\mathbf{X}$ into $\mathbf{Y}_{pred}$.
    }
	\label{fig:anomaly_vs_error}
\end{figure*}

\begin{figure*}[t]
    \centering
    \includegraphics[width=\textwidth]{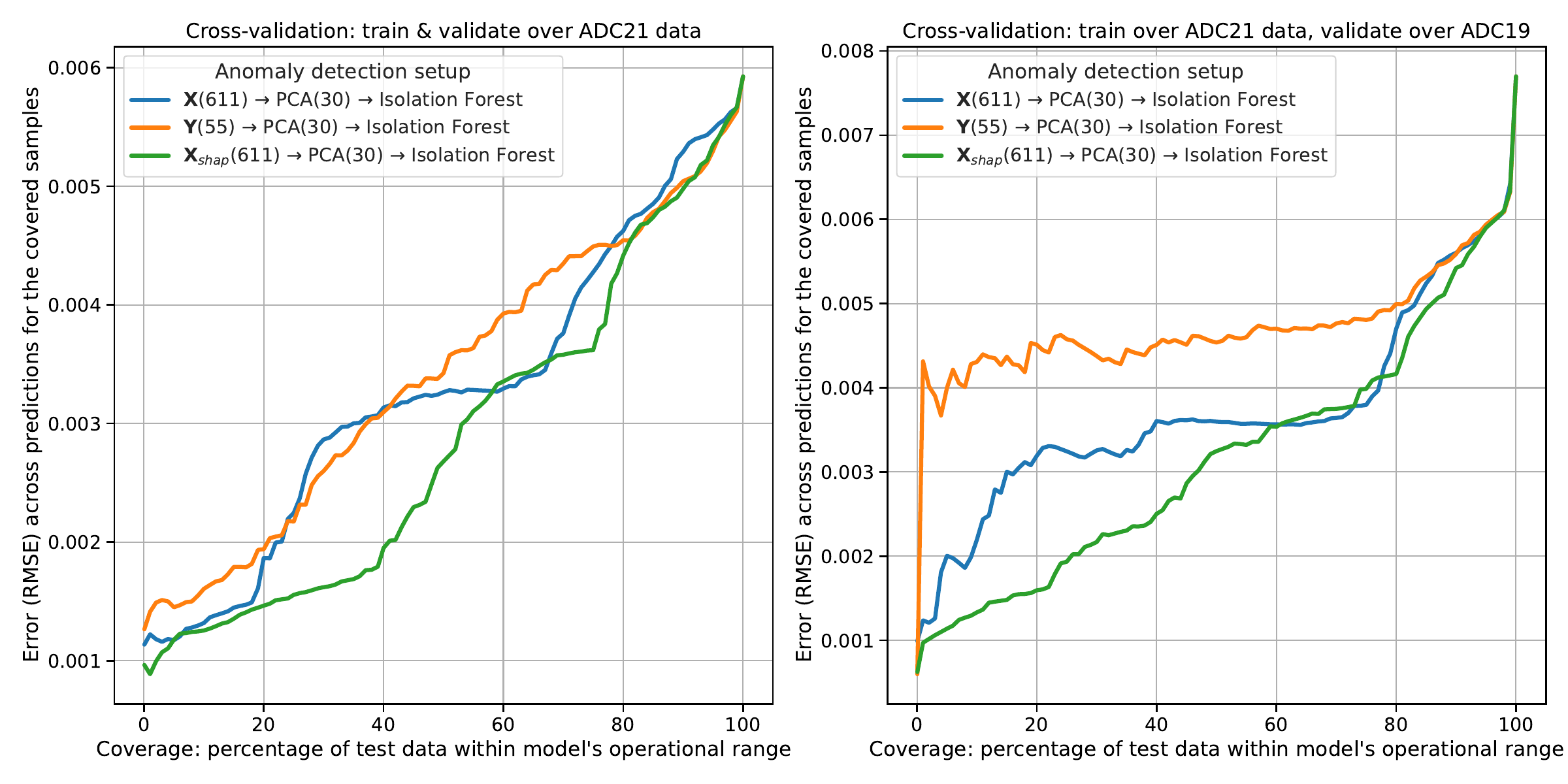}
	\caption{
    Trade-offs achievable between model coverage and predictive performance when acceptance thresholds are defined over Isolation Forests' anomaly scores.
    Averages across the trade-offs seen in individual validation folds.
    }
	\label{fig:coverage_vs_error}
\end{figure*}

\textbf{Datasets.}
We conduct our experiments over the datasets from the 2019 and 2021 Ariel Data Challenges, hereby designated as ADC19 and ADC21 \cite{ADC2019paper}.
The 2019 challenge focused solely on stellar noise in the simulations but assumed the instrument was perfect. In 2021 the signal was convolved with the non-linear and time-dependent instrument response models, introducing detector persistence and non-Gaussian red-noise effects.
From a modelling perspective, those differences enable the assessment of model performance under conditions of data drift (also known as covariate shift) and concept drift \cite{BAYRAM2022108632}.
We use the train data that was made publicly available for those competitions. They contain simulations of 1468 and 1256 star/planet systems, respectively. Experiments across datasets are restricted to a core of 851 systems that were simulated in both datasets, with the same star/planet parameters (\cref{fig:light_curves}).
The data generation process simulated 10 different stellar spot configurations per system, which pollute the signal and therefore have to be identified and corrected for in the derivation of the spectra. Each stellar spot configuration was further corrupted by 10 different instances of additive Gaussian photon noise.

\noindent
\textbf{Data pre-processing pipeline.}
We treat the photon noise instances as repeated observations of the same system, and aggregate data across them.
We make use of the fact that the known planet parameters allow for alignment of independent observations' time series, leaving us in this case with an expectation that the maximum dip in the transit light curve will occur around the $t=150$ time-step. We implement a "mirror-stacking" approach where the observation matrix is stacked with a time-inverted copy of itself, thus exploiting the light curve's expected symmetry (at time $t=140$, for instance, we'll have as well the observations taken at $t=160$).
A regular grid of 11 points is taken across the time series, every 5 time steps between $t=100$ and $t=150$. Per grid point, an aggregation is performed using a harmonic mean across the observations in a time window of radius $r=10$, and across the same windows of the 10 photon noise instances. Each encoded value then results from aggregating across $(r \cdot 2 + 1) \cdot 2 \cdot 10 = 420$ observations.
At this point, data is transformed from relative flux values into relative radii, using the $R_p/R_s = \sqrt{1 - min(1, f)}$ relation.
This has the advantage of simplifying the modelling task, as both model inputs and outputs will now be in the same space.
This process is repeated across observations' 55 wavelength channels, which leads to a $11 \cdot 55 = 605$ dimensional data encoding.
Finally, we stack onto this representation the 6 known star and planet parameters (stellar temperature, surface gravity, etc.).
By following this process, the raw data is transformed into a matrix $\mathbf{X}$ of 12560 611-dimensional samples, in the case of ADC21 data, or 8510 samples in experiments across datasets.
% maybe higlight this is therefore the level at which performance measures are taken (not reporting metrics at the original shape of 125600 samples, neither performing a final aggregation at planet level across predictions for the planet's different stellar spot configurations)

\vspace{5pt}
\noindent
\textbf{Modelling pipeline.}
For the sake of simplicity, and to obtain models having abundant failure contexts, we modelled the problem of extracting transmission spectra from the previously described data using Ridge regression (linear least squares with $L_2$ regularisation).

The modelling of data distributions for the purpose of anomaly detection used Isolation Forest models \cite{IsolationForest:08, IsolationForest_anomaly:12}. These models allow for identification of samples that belong to low density regions of the training data. Scarcity of data in such regions leads one to assume that the prediction model (Ridge in this case) won't have been sufficiently informed as to how to model the target, and its performance will then be expected to be worse there. Isolation Forests produce anomaly scores for the data samples they are given. Values above 0.0 are judged to be inliers with respect to the training data, and values below that outliers.

As we are dealing with high-dimensional datasets, Isolation Forests modelled projections of the data into 30-dimensional spaces.
The projections were obtained using Principal Component Analysis (PCA).

We assess the effectiveness of three anomaly detection setups, which differ in the data that Isolation Forests, and PCA, were asked to model (\cref{fig:anomaly_vs_error}):
\begin{itemize}
    \item $\mathbf{X}$: models the training data that the prediction model takes as input (light curve encodings here).
    
    \vspace{-2pt}
    \item $\mathbf{Y}_{true}$: this setup ignores the inputs, and models the targets instead (transmission spectra here).
    We want to detect when a prediction model generates outputs unlike anything that exists in the training data, as they will likely be wrong.
    
    \vspace{-2pt}
    \item $\mathbf{X}_{shap}$: this setup models the distribution of SHAP values ("SHapley Additive exPlanations") produced by the SHAP Explainable AI library \cite{SHAP} that represent how the prediction model transforms $\mathbf{X}$ into $\mathbf{Y}_{pred}$.
\end{itemize}

A challenge in tying anomaly scores to another model's predictive performance lies in that the anomaly detection model lacks sensitivity to the kinds of data shifts that most impact the other model.
The prediction model values input dimensions differently depending on their usefulness to generate the prediction. To an anomaly detection model, in contrast, all dimensions are equally important.
The $\mathbf{X}_{shap}$ setup addresses this limitation by modelling values that blend $\mathbf{X}$ and the internal structure of a prediction model trained to obtain $\mathbf{Y}_{true}$ from it.
The Ridge model's SHAP values were calculated using a \texttt{LinearExplainer} with interventional feature perturbation. % (default setting).
In this approach, SHAP values are given by $\phi_i = \beta_i \cdot (x_i - \mu_i)$, where $\beta_i$ represents the coefficient of the $i$-th feature in the linear model, $x_i$ is the value of that feature for the instance being explained, and $\mu_i$ is the feature's mean value across the train set.
This being a multi-output regression problem, we obtain 55 vectors of 611 SHAP values per input sample.
The median SHAP is calculated for each input variable to obtain the sample's 611-dimensional vector used by PCA and, after projection, by the Isolation Forest.

We used the Ridge, PCA and Isolation Forest implementations available in the \texttt{scikit-learn} 0.24.2 library \cite{scikit-learn}, with their default parameter settings.
% https://scikit-learn.org/0.24/
% https://scikit-learn.org/0.24/modules/generated/sklearn.linear_model.Ridge.html
% https://scikit-learn.org/0.24/modules/generated/sklearn.decomposition.PCA.html
% https://scikit-learn.org/0.24/modules/generated/sklearn.ensemble.IsolationForest.html
%
SHAP values were obtained with the \texttt{shap} 0.39.0 library \cite{SHAP}.
% \code{feature\_perturbation='interventional'}
% https://shap-lrjball.readthedocs.io/en/latest/generated/shap.LinearExplainer.html

\vspace{5pt}
\noindent
\textbf{Cross-validation setup.}
Experiments carried out using Group 10-fold cross-validation.
The data pre-processing described above results in datasets with groups of 10 samples belonging to the same planet, having different stellar spot noise configurations. Cross-validation then ensured that samples belonging to the same planet would always be together in either a train- or validation-fold.
Experiments across the ADC19 and ADC21 datasets, having fewer planets, used instead 5-fold CV to increase the size and diversity of validation folds. In those experiments, planets in the train-fold were modelled using encodings of their ADC21 data, but planets placed in validation-folds used instead encodings of their ADC19 data. This allows us to assess effectiveness of the proposed approaches in contexts of degraded model performance due to drift.

Data standardisation into zero mean and unit variance was implemented, as these are important pre-processing steps for Ridge and PCA. Standardisation was defined over each train-fold, and then applied there and in the corresponding validation fold.

Predictive performance was assessed using the Root Mean Squared Error (RMSE) metric, chosen for its sensitivity to large errors, thus expressing the preference for all values in a spectrum to be accurately predicted.
% It was measured at two levels: in \cref{fig:anomaly_vs_error} it is taken across the 55 wavelength values of a sample, and in \cref{fig:coverage_vs_error} it is taken across all predictions for samples deemed to be in the model's operational range, after applying an acceptance threshold based on anomaly scores.

\vspace{5pt}
\noindent
\textbf{Results.}
With the previously presented setup, Ridge models achieve a RMSE of 0.006085 when cross-validating across the full ADC21 dataset, and a degradation to 0.008421 when cross-validation trains over ADC21 and validates over ADC19.
% In comparison, the Mean Absolute Error is of 0.002999 and 0.003410 respectively across those setups.
For context: training over the full ADC21 dataset and evaluating over the full ADC19 dataset leads to a RMSE of 0.008589 (0.010306 in the other direction).
% ^ NOTE: here "full" means the full trainsets under consideration, which are actually just the train sets shared publicly in the competitions, as previously noted.
The targets for the 851 considered planets have differences across both datasets that evaluate to a RMSE of 0.013431.

\cref{fig:anomaly_vs_error} shows the anomaly scores measured across the validation folds having a moderate Spearman's rank correlation with the RMSE values taken over the sample's 55 wavelength channels.
The strongest correlation is observed in the $\mathbf{X}_{shap}$ setup, which validates the hypothesis that aligning the modelled distribution with the behaviour of the prediction model improves the capacity to identify performance degradation.

\cref{fig:coverage_vs_error} shows the effectiveness of a safety cage architecture that delimits the prediction model's operational range by imposing acceptance thresholds over the anomaly scores obtained through the approaches presented here.
It shows the coverage \textit{vs.} error trade-offs achievable as that threshold varies across the span of anomaly scores observed in the validation folds.
Again we see the $\mathbf{X}_{shap}$ setup as being the most effective one. It presents the slowest curve growth, which corresponds to an operational range boundary that is more accurate at including within it the performant regions, and leaving outside the error-prone ones.
The right plot shows the transition to ADC19 data resulting in a large performance degradation, but mostly due to large errors in samples that are easily flagged by the Isolation Forest.
An acceptance threshold for anomaly scores $\ge -0.071$ would reduce coverage by $\thicksim 2\%$ and RMSE by $\thicksim 20\%$.
%
% Note: the error levels for 100% coverage (top right) are lower than the CV RMSE values reported above. That's due to the lines on the plot being averages of the lines for each individual validation-fold, whereas the CV RMSE values reported here result from measuring over all the predictions obtained across the validation-folds.

For acceptance thresholds that reduce coverage by at most 20\%, results indicate that the three setups achieve identical performance levels. Since the calculation of SHAP values can be computationally costly for certain types of models, the $\mathbf{X}$ and $\mathbf{Y}_{true}$ setups may be preferable in scenarios where coverage is prioritised over predictive performance.

\section{Discussion}

In missions like Ariel, prediction failures by Machine Learning models pose risks, but also opportunities.
Data products obtained through ML-assisted processing of the mission's data cannot be contaminated by wrong extrapolations beyond models' training data. The architecture explored here succeeds at providing safeguards against such risks.
We have tackled the problem of identifying out-of-domain samples, which enables us to recognise contexts of high epistemic uncertainty. As future work we plan to extend the approach with uncertainty quantification methods, which will provide identification as well for contexts of high aleatoric uncertainty \cite{PSAROS2023111902}.

Beyond such risks, there are also opportunities.
Anomaly/novelty detection may point to observation targets of particular scientific interest, due to their uniqueness \cite{Forestano2023}.
Current modelling failures are not guaranteed to remain as failures if subsequent effort is allocated to better simulate and understand the failure cases.
Some failures result from sparsity in the training data that doesn't allow the modelling process to capture the patterns of how to process such data. This sort of feedback is valuable to the effort of gradually building the best possible simulations of the targets of interest, which improves models' training datasets.
The gaps in coverage left by delimiting a model's operational range may also be filled by different models, potentially of different kinds, that specialise at accurate prediction where previous models fail. This architecture may also assist such efforts.

\subsection*{Acknowledgements}
Research funded by the ESA Science Faculty Research projects "Machine Learning Quality Assurance in Ariel and beyond" and "Ariel Machine Learning Data".

\printbibliography
\addcontentsline{toc}{section}{References}

\end{document}